\documentclass{article}
\usepackage{arxiv}

\usepackage[utf8]{inputenc} % allow utf-8 input
\usepackage[T1]{fontenc}    % use 8-bit T1 fonts
\usepackage{palatino}
\usepackage{hyperref}       % hyperlinks
\usepackage{url}            % simple URL typesetting
\usepackage{booktabs}       % professional-quality tables
\usepackage{amsfonts}       % blackboard math symbols
\usepackage{nicefrac}       % compact symbols for 1/2, etc.
\usepackage{microtype}      % microtypography
\usepackage{xcolor}         % colors

\definecolor{darkblue}{rgb}{0, 0, 0.5}
\hypersetup{
    colorlinks=true,
    citecolor=darkblue,
    linkcolor=darkblue,
    urlcolor=darkblue,
}

\usepackage[verbose=true,letterpaper]{geometry}
\usepackage[sort,numbers]{natbib}

\usepackage[hyphenbreaks]{breakurl} %for arXiv's idiotic hyperref!
\usepackage{microtype}
\usepackage{hyperref}       % hyperlink
\usepackage{url}            % simple URL typesetting
\usepackage{booktabs}       % professional-quality tables
\usepackage{amsfonts}       % blackboard math symbols
\usepackage{nicefrac}       % compact symbols for 1/2, etc.
\usepackage{microtype}      % microtypography
\usepackage{amsmath,bm,mathtools} 
\usepackage{float}
\usepackage{amssymb}
\usepackage{amsthm}
\usepackage{xspace}
\usepackage{enumerate}
\usepackage{graphicx}
\usepackage{pgf}
\usepackage{paralist}
\usepackage{wrapfig}
\usepackage{nicefrac}       % compact symbols for 1/2, etc.

\setcitestyle{authoryear,open={(},close={)}}

\usepackage{fullpage}
\usepackage{amsmath,amsfonts,bm}

\def\eqref#1{equation~\ref{#1}}
\def\1{\bm{1}}

\DeclareMathAlphabet{\mathsfit}{\encodingdefault}{\sfdefault}{m}{sl}
\SetMathAlphabet{\mathsfit}{bold}{\encodingdefault}{\sfdefault}{bx}{n}

\usepackage{url}
\usepackage{graphicx}
\usepackage{wrapfig}
\usepackage{bbm}
\usepackage{amsmath}
\usepackage{amssymb}
\usepackage{amsthm}
\usepackage{natbib}
\usepackage{xcolor}
\usepackage{multirow}
\usepackage{setspace}
\usepackage{enumitem}
\usepackage{cleveref}

\theoremstyle{plain}

\usepackage{listings}
\usepackage{siunitx}
\usepackage{bookmark}
\usepackage{textcomp}
\usepackage{multirow}
\usepackage{nth}
\usepackage{indentfirst}
\usepackage{siunitx}
\usepackage{changepage}
\usepackage{bm}
\usepackage{setspace}
\usepackage{natbib}
\usepackage{multirow}
\usepackage{multicol}
\usepackage{color, colortbl}
\usepackage{booktabs}
\usepackage{amsmath}
\usepackage{amsthm}
\theoremstyle{remark}

\usepackage{subfig}
\usepackage{wrapfig}
\usepackage{enumitem}
\usepackage{makecell}
\usepackage[dvipsnames, svgnames, x11names]{xcolor}
\definecolor{darkmagenta}{rgb}{0.56, 0.0, 1.0}  

\hypersetup{colorlinks,linkcolor={blue},citecolor={darkmagenta},urlcolor={blue}}
\usepackage[most,skins,theorems,breakable]{tcolorbox}
\tcbset{
  mybox/.style={
    width=\linewidth,
    top=14pt,
    bottom=7pt,
    colback=cyan!5!white,
    colframe=black,
    colbacktitle=black,
    enhanced,
    center,
    breakable,
    attach boxed title to top left={yshift=-0.13in,xshift=0.15in},
    boxed title style={boxrule=0pt,colframe=white,},
  }
}
\newtcolorbox{mybox}[2][]{mybox,title=#2,#1}

\usepackage{titling}
\usepackage{setspace}

\usepackage{algorithm}
\usepackage{algorithmicx}
\usepackage{algpseudocode}

\colorlet{softred}{red!80!black}
\colorlet{softgreen}{green!55!black}
\colorlet{softblue}{blue!65!black}
\colorlet{softyellow}{yellow!70!black}
\colorlet{softgray}{black!50}
\definecolor{softblue}{RGB}{38, 139, 210}

\usepackage{array}
\newcolumntype{L}[1]{>{\raggedright\arraybackslash}p{#1}}
\newcolumntype{C}[1]{>{\centering\arraybackslash}p{#1}}

\newcommand\blfootnote[1]{%
  \begingroup
  \renewcommand\thefootnote{}\footnote{#1}%
  \addtocounter{footnote}{-1}%
  \endgroup
}

\title{\textbf{Local Guidance, Global Impact: Gaussian-Reshaped \\ Trust Region Unlocks Behavior Transitions}}
\author{
{{\textbf{Bingxu Liu$^{1,4*}$ ~ Jiashun Liu$^{1*}$ ~ Johan Obando-Ceron$^{2,3}$ ~ Hao Wang$^{5}$ ~ Runze Liu$^{1}$}}}\\
{{\textbf{Pablo Samuel Castro$^{2,3}$ ~ Aaron Courville$^{2,3}$ ~ Ling Pan$^{1\dagger}$}}}\\
{\normalsize{$^{1}$Hong Kong University of Science and Technology $^{2}$Mila - Qu\'ebec AI Institute}}\\
{\normalsize{$^{3}$Universit\'e de Montr\'eal $^{4}$Fudan University $^{5}$City University of Hong Kong}}\\
}
\date{}

\begin{document}

\maketitle

\vspace{-3.0em}
\blfootnote{$^*$ Equal contribution\quad$^\dagger$ Correspondence to \texttt{lingpan@ust.hk} }

\begin{abstract}
While Proximal Policy Optimization (PPO) demonstrates strong performance in stationary settings, we show that its standard optimization paradigm struggles in continual and non-stationary environments. The failure does not stem from insufficient model capacity or overly restrictive clipping. Instead, PPO performs persistent, directionally inefficient local updates, which indicates a lack of geometry-aware guidance for accumulating meaningful behavioral change and ultimately hindering transitions toward new behavior patterns. Although divergence-based regularization introduces partial geometric awareness, its monotonically increasing penalties implicitly discourage large policy deviations, even when such shifts are necessary for effective adaptation. To address this limitation, we propose \textbf{Gaussian Trust Region Policy Optimization} (\texttt{GTR}), which reshapes the trust region using a Gaussian kernel. The resulting constraint is bounded and non-monotonic, providing strong local stability while progressively relaxing under sustained high-advantage updates. To further improve robustness, we introduce a \emph{Mixture Gaussian Anchor} that adapts to recent policy trajectories, reducing variance induced by stale references. \texttt{GTR} is architecture-agnostic and achieves strong performance across games, simulated robotic control, open-world exploration, and language model post-training. These results demonstrate that geometry-aware trust-region design can be a promising direction for robust reinforcement learning in complex non-stationary environments. \textbf{Our code is available \href{https://anonymous.4open.science/r/GTR_demo/README.md}{here.}}
\end{abstract}

\section{Introduction}\label{intro}
Proximal Policy Optimization (PPO)~\citep{schulman2017proximal}, a standard deep reinforcement learning (RL) algorithm \citep{sutton1998reinforcement}, has achieved significant success across diverse domains, including robotic control~\citep{hwangbo2019learning,akkaya2019solving}, game AI~\citep{berner2019dota}, and large foundation model post-training~\citep{ouyang2022training,guo2025deepseek}. Its success largely stems from a simple yet effective principle: constraining policy updates to remain close to a reference policy via a clipping mechanism \citep{schulman2017proximal}, ensuring stable improvement under advantage-driven optimization \citep{kakade2002approximately,schulman2015trust}. This local-update paradigm is well-suited to traditional, stationary environments, where progress can often be made through incremental refinement around the current behavioral mode~\citep{andrychowicz2020matters}.

\begin{wrapfigure}{r}{0.58\textwidth}
\centering
%\vspace{-0.28cm}
    \includegraphics[width=\linewidth]{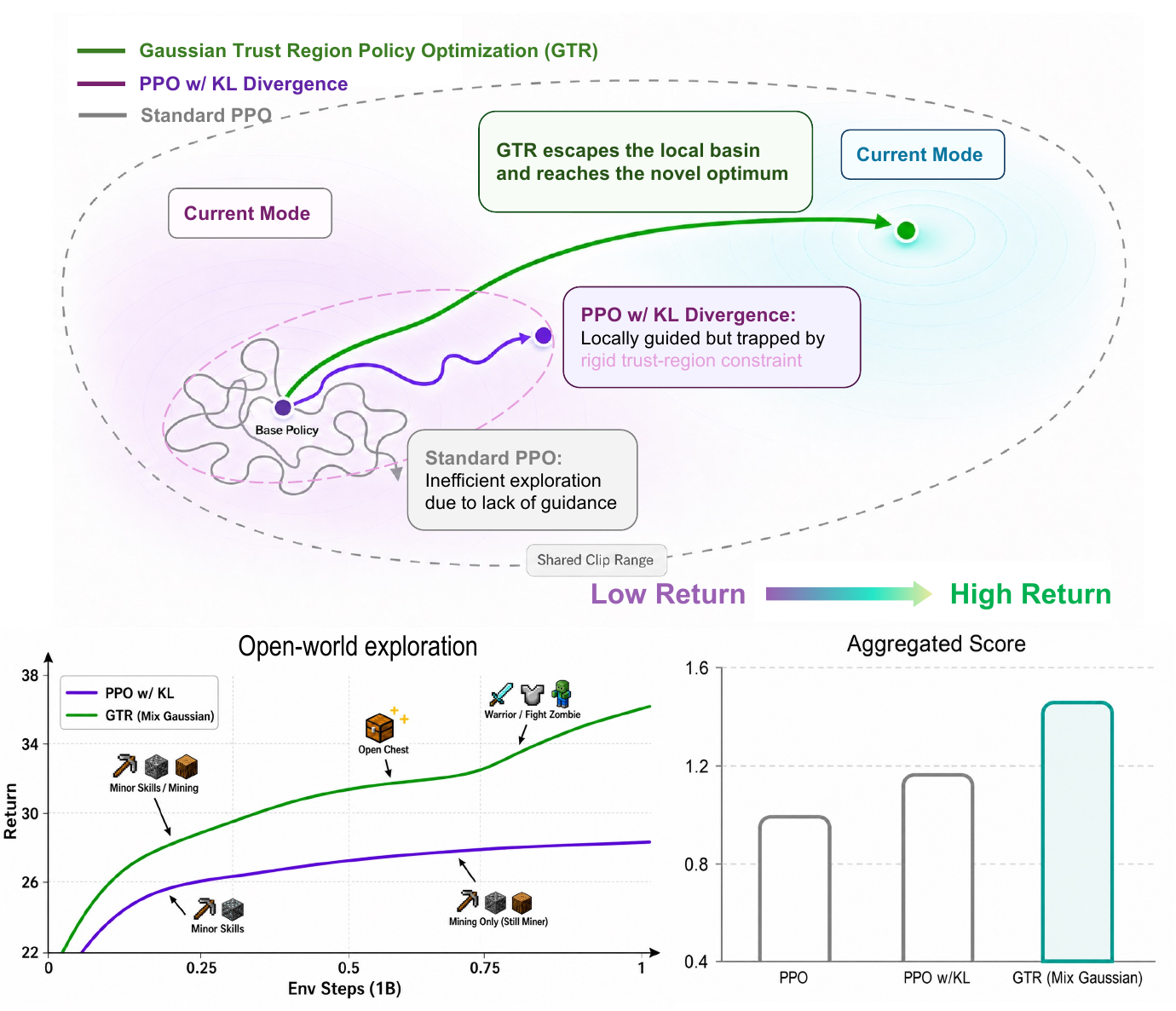}
    \vspace{-0.6cm}
    \caption{(Top): \textbf{In continual learning, standard PPO may perform persistent but ineffective local policy search and fail to accumulate the distributional shift required to reach a better behavioral mode.} Monotone divergence regularization provides local guidance but cannot adapt to a different mode stably due to the monotonically increasing penalty. (Bottom): \textbf{\texttt{GTR} achieves continuous mode transition in open-world and achieves the best performance in all benchmarks.}}
    \label{fig:overview}
    \vspace{-0.3cm}
\end{wrapfigure}
However, as modern RL systems move toward continual, non-stationary regimes~\citep{wolczyk2021continual,khetarpal2022towards} with evolving dynamics, this paradigm requires rethinking in the context of continual RL \citep{abel2023definition,willi2024mixtureexpertsmixturerl,tang2025mitigating,pan2025survey}. When tasks or required skills evolve over time, effective learning requires not only stable local improvement, but also the ability to leave outdated behavioral modes and transition toward newly advantageous ones~\citep{agarwal2021theory}. 
In such dynamic scenarios, the clipped objective may provide insufficient geometric 
preference among competing update directions, and we observe that standard PPO fails to achieve such effective transitions. As a result, the policy can undergo persistent but poorly directed local optimization and eventual collapse, as illustrated in \autoref{fig:overview} (Top). This failure highlights a fundamental limitation of standard PPO in continual learning: \emph{it lacks a mechanism for shaping local updates according to the distributional geometry induced by the policy, making advantage-driven optimization prone to unproductive drift rather than meaningful behavioral transition.
}

Intuitively, effective guidance should shape the local optimization geometry \citep{kakade2001natural} for distinguishing directions based on how sensitively policy distributions respond to parameter perturbations. This can be characterized by the Fisher Information Matrix (FIM): high-curvature directions induce large distributional shifts, while low-curvature directions correspond to relatively stable updates \citep{amari1998natural}. Divergence-based regularization introduces partial landscape awareness through its connection to the FIM, e.g., the second-order expansion of KL divergence recovers the Fisher metric \citep{schulman2015trust}. Empirically, we find that incorporating divergence-style penalties improves stability and alleviates inefficient exploration, suggesting that the absence of geometric guidance is a key source of adaptation failure in standard PPO. Nevertheless, most divergence penalties increase monotonically with policy deviation, implicitly assuming that larger shifts are undesirable \citep{nowozin2016f}. While reasonable in stationary settings, this assumption becomes restrictive in continual RL, where optimal behaviors may lie far from the current policy \citep{lecarpentier2021lipschitz}. Therefore, overly conservative regularization can hinder necessary mode transitions as shown in \autoref{fig:overview} (Top), raising a key question: {\textbf{
can we design a trust-region regularization that is stiff enough to suppress noisy local drift, yet permissive enough to allow sustained high-advantage updates to escape the current behavioral mode?}}

Idealized trust-region regularization should enforce stiffness around small deviations to stabilize incremental learning, while gradually relaxing as the policy is consistently driven toward high-advantage situations. This balance prevents unstable local updates while avoiding over-constraining exploration when strong advantage signals suggest departure from suboptimal behaviors. Guided by this principle, we introduce a Gaussian-shaped trust-region constraint, which induces localized stiffness near the reference policy and a rapidly diminishing penalty for larger deviations. This design also encodes a geometry-aware regularization: the Gaussian weighting modulates the influence of policy ratios in a distance-dependent manner, effectively reweighting gradient contributions and shaping the local optimization landscape. In contrast to monotonic divergence penalties, this Gaussian penalty allows the policy to preserve stability in the extreme proximal region while reducing resistance to sustained, high-advantage updates, thereby facilitating controlled exploration. However, as the reference policy becomes stale under a high-data-reuse setting, a fixed Gaussian anchor can be unstable. Therefore, we propose \emph{Mixture Gaussian Anchor}, which models the constraint as a dynamically updated mixture over the recent policy snapshots, to reduce variance induced by outdated references while remaining responsive to recent high-advantage updates. We instantiate this framework as \textbf{Gaussian Trust Region Policy Optimization} (\texttt{GTR}).

\texttt{GTR} improves adaptability at the algorithmic level and generalizes seamlessly across diverse policy classes, i.e., MLPs \citep{mnih2015human,schwarzer23a,ceron24b}, RNNs \citep{chung2014empirical}, SimBa \citep{lee2024simba}, and Transformer-based \citep{vaswani2017attention} policies. Extensive evaluations on robotic control \citep{todorov2012mujoco,tassa2018deepmind}, open-world exploration \citep{matthews2024craftax}, and language model post-training \citep{yu2025dapo,guo2025deepseek} demonstrate that \texttt{GTR} consistently achieves state-of-the-art performance, as shown in \autoref{fig:overview}, highlighting that principled constraint design could be a promising way for policy improvement. Our contributions are threefold:
\definecolor{blueviolet}{RGB}{138,43,226}
\newtcolorbox{insightblock}{
  colback=blueviolet!5,   
  colframe=blueviolet!50!black!50!,    
  boxrule=0.5mm,       
  arc=2mm,            
  left=0pt,           
  right=8pt,           
  top=8pt,            
  bottom=8pt,}
\begin{insightblock}
\begin{enumerate}[leftmargin=1.5em]
    \item We reveal that a key factor for the collapse of PPO in continual RL is that it exhibits persistent but directionally inefficient local updates, which indicates a lack of local optimization geometry.
    \item We introduce \texttt{GTR}, a Gaussian-shaped, geometry-aware trust-region regularization that balances local stability and global adaptability in continual setups.
    \item We extensively validate \texttt{GTR} across a wide range of benchmarks and various architectures. \texttt{GTR} consistently improves performance in non-stationary and continual learning settings.
\end{enumerate}
\end{insightblock}

\section{Preliminaries}\label{sec:pre}
\paragraph{Proximal Policy Optimization} Standard Proximal Policy Optimization (PPO)~\citep{schulman2017proximal} is a policy gradient method that improves training stability by constraining policy updates within a proximal region of a reference policy. Let $\pi_\theta$ denote the current policy and reference policy $\pi_{\theta_{\mathrm{old}}}$ used for data collection. PPO optimizes a surrogate objective based on importance sampling:$
r_t(\theta) = \frac{\pi_\theta(a_t \mid s_t)}{\pi_{\theta_{\mathrm{old}}}(a_t \mid s_t)},
$where $r_t(\theta)$ is the likelihood ratio. The clipped objective is defined as:$
L^{\mathrm{CLIP}}(\theta) = \mathbb{E}_t \left[
\min \big(
r_t(\theta) \hat{A}_t,
\mathrm{clip}(r_t(\theta), 1 - \epsilon, 1 + \epsilon) \hat{A}_t
\big)
\right],
$where $\hat{A}_t$ is an estimate of the advantage function and $\epsilon$ is a hyperparameter controlling the trust region size. However, the clipping mechanism does not explicitly account for the geometry of the policy space, as it constrains updates based solely on the likelihood ratio without incorporating curvature information.
\paragraph{Trust region regularization can be used as directional guidance} We consider a regularized policy optimization objective of the standard form \citep{schulman2015trust}:
\begin{equation}
\max_\theta \;\;
\mathcal{L}(\theta)
= \mathbb{E}\big[r_t(\theta) \hat{A}_t\big]
- \lambda \, D(\pi_{\theta_{\mathrm{old}}} \,\|\, \pi_\theta),
\end{equation}
where $D(\cdot)$ is a divergence measure. Under a small update $\theta = \theta_{\mathrm{old}} + \Delta \theta$, if $D$ is the KL divergence, the objective admits a second-order approximation:
\begin{equation}
\mathcal{L}(\theta) \approx
g^\top \Delta \theta
- \tfrac{\lambda}{2} \Delta \theta^\top F \Delta \theta,
\end{equation}
where$
g = \mathbb{E}\big[\hat{A}_t \nabla_\theta \log \pi_\theta(a_t|s_t)\big],
\quad
F = \mathbb{E}\big[\nabla_\theta \log \pi_\theta \nabla_\theta \log \pi_\theta^\top\big]
$is the Fisher Information Matrix (FIM). The optimal update is given by:
$
\Delta \theta^* = \frac{1}{\lambda} F^{-1} g.
$
Compared to standard policy gradients $\Delta \theta \propto g$, trust-region regularization transforms the update into a geometry-aware form $F^{-1}g$, where the FIM induces anisotropic scaling across directions. In particular, directions associated with high curvature (large eigenvalues of $F$) are suppressed, while low-curvature directions are amplified. 

This shows that trust-region regularization acts as a \emph{directional filter}, guiding updates based on the local geometry of the policy space rather than purely on scalar advantage signals.

\section{Trust Region Regularization Unlocks Behavior Transition}\label{sec:2.1}
\begin{wrapfigure}{r}{0.5\textwidth}
\centering
\vspace{-0.5cm}
    \includegraphics[width=\linewidth]{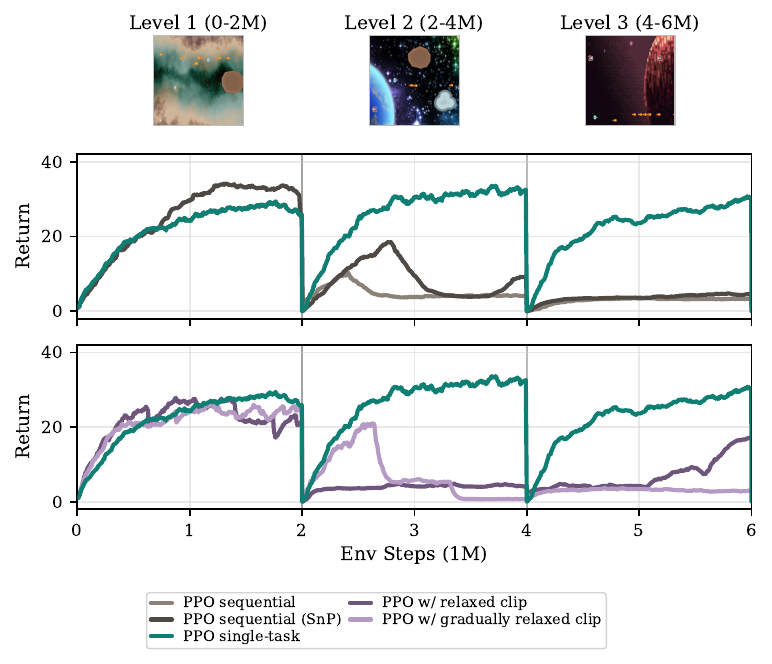}
    \vspace{-0.4cm}
    \caption{(Top): Complex layouts across levels. (Middle): PPO collapses during sequential training, even w/ repairing the network. (Bottom): With higher clip range, PPO still collapses.}
    \label{case_study}
    \vspace{-0.2cm}
\end{wrapfigure}\paragraph{Network capacity is not the bottleneck.}
To test whether the failure is caused by limited model expressiveness, we incorporate stochastic network perturbations (SnP)~\citep{ash2020warm} to continually refresh network capacity (\textcolor{darkgray}{PPO-sequential-SnP}) (\autoref{case_study} (Middle)). Despite improved performance on the first task, the policy still collapses, suggesting that insufficient capacity is not the primary cause. We first investigate the root cause of PPO’s limited behavioral pattern transition capability using a controlled sequential training setup inspired by \citep{muppidi2024pick, abbas2023loss}. Specifically, we construct a curriculum on the challenging starpilot environment from Procgen~\citep{cobbe2020leveraging}, where task difficulty increases progressively (Level 1$\to$Level 2$\to$Level 3). We train a single MLP policy using standard PPO across the full sequence (\textcolor{gray}{PPO-sequential}), and compare it to a control setting where PPO is trained from scratch on each task independently (\textcolor{teal}{PPO-single task}). As shown in \autoref{case_study} (Bottom), PPO successfully solves each task in isolation. However, under sequential training, although the policy rapidly learns the first task coincides with \textcolor{teal}{PPO-single task}, it experiences a sharp performance collapse when transitioning to subsequent tasks. This indicates that the failure is not due to task difficulty alone, but arises from the interaction between PPO’s update mechanism and non-stationary task shifts.

\paragraph{Larger clipping ranges fail to recover the policy.}
A natural hypothesis is that the clipping constraint restricts the policy to a region too small to capture new behaviors \citep{wang2019trust}. To evaluate this, we relax the clipping bound (\textcolor{violet}{PPO w/ relaxed clip}) and also consider schedules where the clipping range grows over time (\textcolor{Plum}{PPO w/ gradually relaxed clip}). Neither modification recovers continual learning ability, indicating that clipping is not the root cause (\autoref{case_study} (Bottom)).

\begin{figure}[h]
\centering
\vspace{-0.2cm}
    \includegraphics[width=\linewidth]{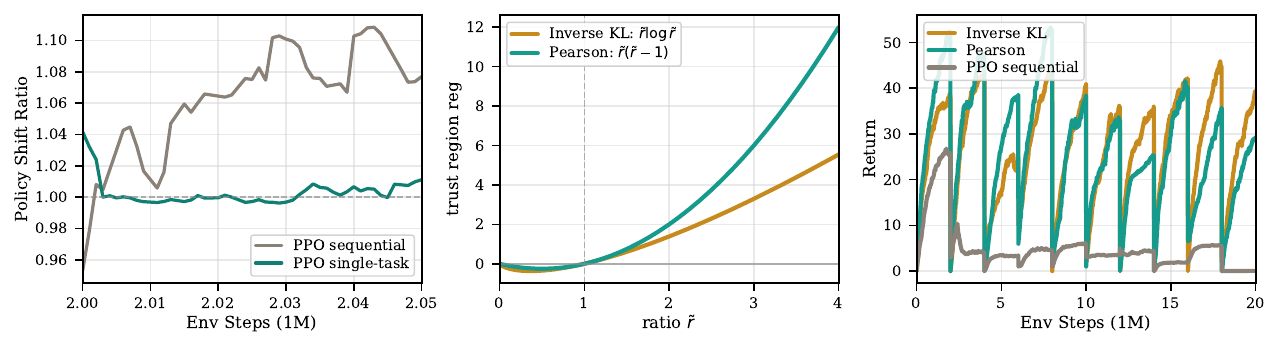}
    \caption{(Left): \textbf{Policy update magnitude of standard PPO is high.} After the task switch, PPO shows a higher update range than the baseline, and the updated policy is always far from the reference policy, which indicates that it cannot sense the reliable optimization direction. (Middle): \textbf{Visualization of constraint strength.} The penalty strength of the divergence corresponding to the shift ratio. (Right): \textbf{Trust-region regularizations unlock the behavior transition ability of PPO.} Under the setup of \autoref{case_study} (Level 1 to Level 10), the policy employing these two mild punishment divergences achieves better continual learning ability compared to the standard PPO.}
    \label{procgen_div}
    \vspace{-0.2cm}
\end{figure}

\paragraph{Failure arises from aimless local optimization.}
We further examine the dynamics preceding collapse. As shown in \autoref{procgen_div} (Left), the policy shift ratio exhibits high-frequency fluctuations before failure, while gradient magnitudes remain significant. This suggests that the policy is actively updating and exploring in parameter space. However, these updates fail to produce consistent improvement, indicating that exploration is directionally inefficient despite being persistent.

\paragraph{Trust-region regularization is critical for effective adaptation.}
The above results indicate that PPO fails in continual settings not due to insufficient model capacity or overly restricted update magnitude, but because it lacks the navigation in complex local optimization landscapes. In particular, the absence of directional guidance leads to inefficient exploration, causing the accumulation of biased local updates and resulting in training collapse. This suggests that the core issue lies in how trust-region regularization shapes the update geometry, rather than in the scale of the update itself.

To test this hypothesis, we introduce divergence-based penalties, which provide partial landscape awareness through their connection to the Fisher Information Matrix (FIM). Unlike standard PPO, which relies solely on scalar advantage weighting, these penalties incorporate curvature information that differentiates update directions based on their induced distributional shifts. As shown in \autoref{procgen_div} (Middle), both \textcolor{brown}{inverse KL}~\citep{tang2025few,shah2025comedy} and \textcolor{DarkGreen}{Pearson $\chi^2$}~\citep{xie2024simple} penalties lead to immediate and consistent performance improvements over \textcolor{gray}{standard PPO} in sequential Procgen training (\autoref{procgen_div}, right). These results provide empirical evidence that incorporating geometry-aware regularization significantly improves local optimization quality. By guiding updates toward directions that yield stable and meaningful policy changes, such regularization enables iterative improvements to accumulate, resulting in gradual behavioral transitions and sustained performance in continual learning.

\section{Reshape the Trust Region via Gaussian Kernel}\label{theointro}
\begin{wrapfigure}{r}{0.48\textwidth}
\centering
\vspace{-0.4cm}
    \includegraphics[width=\linewidth]{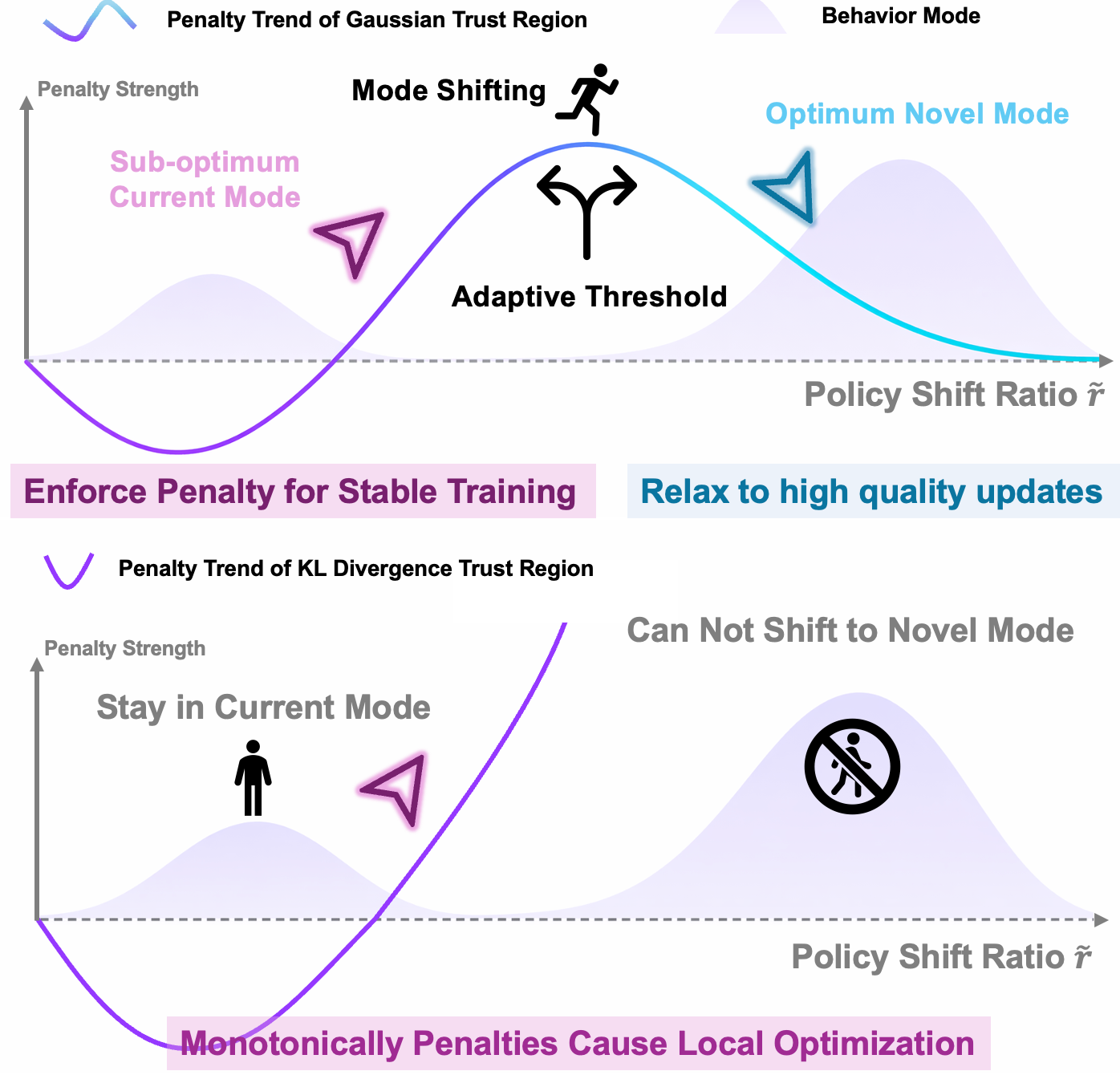}
    \caption{\textbf{Visualization of constraints.} Compared to KL divergence, Gaussian maintains proximal stability while allowing far exploration driven by high advantage.}
    \label{starpilotgaussian}
\vspace{-0.3cm}
\end{wrapfigure}
As \autoref{sec:2.1} shows, incorporating divergence-based penalties can achieve behavior transition by providing geometry-aware guidance for local optimization. However, these approaches remain fundamentally limited. Most divergence-based penalties increase monotonically with policy deviation, implicitly enforcing a linear or unbounded growth of constraint strength as the policy moves away from the reference. In non-stationary environments, effective adaptation often requires transitioning toward behavior modes that are substantially different from the current policy. In such cases, monotonically increasing penalties can overly suppress exploration precisely when large, consistent updates are needed. This observation suggests that an effective trust-region constraint for continual learning should not only encode local geometry, but also relax its restriction in a controlled manner as the policy moves toward promising regions. In other words, the regularization should provide strong local stability while avoiding excessive resistance to sustained, high-quality updates far away from the current reference policy.

To address this limitation, we propose to reshape the trust region using a Gaussian kernel. This formulation induces a bounded, non-monotonic constraint that preserves local geometry while allowing controlled exploration beyond the immediate neighborhood, enabling efficient navigation of complex optimization landscapes: (i) \textbf{Forgetting Mechanism:} $\lambda(\tilde{r}) \rightarrow 0$ as $\tilde{r} \rightarrow 0$. When an action is no longer sampled, no regularization gradient should be applied to it. (ii) \textbf{Unchanged Policy:} $\lambda(\tilde{r}) \rightarrow 0$ as $\tilde{r} \rightarrow 1$. No regularization is enforced when the policy remains unchanged. (iii) \textbf{Local Trust Region:} $\frac{\lambda(\tilde{r})}{\tilde{r}} \approx \pm \epsilon$ when $\tilde{r} = 1 \pm \epsilon$, preserving first-order trust-region behavior and endowing the regularization term with the functionality of a trust region, analogous to the $\epsilon$-alignment \citep{xie2024simple}. (iv) \textbf{Vanishing Constraint at Scale:} $\lambda(\tilde{r}) \rightarrow 0$ as $\tilde{r} \gg 1$, allowing sustained, high-advantage updates to escape local modes. The complete derivation is deferred to \autoref{derivegaussian}.

Guided by these principles, we introduce a Gaussian decay kernel (as shown in \autoref{starpilotgaussian} (Top)), yielding $\lambda_\mathrm{Gaussian}(\tilde{r}) = \tilde{r}(\tilde{r}-1)e^{-(\tilde{r}-1)^2}$, which satisfies localized stiffness near $\tilde{r}\approx 1$ while inducing rapidly vanishing gradients for large deviations. This behavior allows policy updates to remain stable in local regions while avoiding excessive penalization of large, high-advantage shifts. Remarkably, this formulation admits an elegant anti-derivative:
\begin{equation}\label{eq.loss}
    L_\mathrm{Gaussian} = \frac{1}{2}(1 - e^{-(\tilde{r}-1)^2})
\end{equation}
This is a plug-and-play module that can be effortlessly integrated into standard PPO. We term this idealized approach which aligns with the theory as \textbf{Gaussian Trust Region Policy Optimization} (\texttt{GTR}), with the objective function formulated as follows:
\begin{equation}
  L_{\mathrm{GTR}} = \mathbb{E}_{t} \left\{ -\min \left[ r_t(\theta) \hat{A}_t, \mathrm{clip}(r_t(\theta), 1 - \epsilon, 1 + \epsilon) \hat{A}_t \right] + \frac{\eta}{2} \left( 1 - e^{-(r_t-1)^2} \right) \right\}
\end{equation}
\paragraph{Sliding Trust Region via Gaussian Mixtures}While the single Gaussian kernel improves the functional shape of the trust region, it does not consider another implicit issue: During multi-epoch updates, the trust-region anchor remains tied to the behavior policy, causing likelihood ratios to become increasingly unstable as the policy drifts away. This issue is further exacerbated under high UTD regimes, where repeated reuse of stale data leads to biased updates and degraded adaptation \citep{gan2024transductive,li20253po}. We reinterpret it as a consequence of a \emph{rigid trust region}. The constraint remains anchored to an outdated policy and fails to account for the progressive shift of the policy distribution throughout optimization. To address this limitation, we propose a \emph{Mixture Gaussian Anchor}, which generalizes the trust-region constraint by replacing the single reference policy with a mixture over recent policy snapshots. Concretely, we compute the trust-region penalty using multiple intermediate policies from different PPO epochs and average their contributions. This yields a trajectory-aware constraint that tracks the progressive shift of the policy distribution, mitigating the effect of stale references while preserving sensitivity to recent updates. Moreover, due to the vanishing tails of the Gaussian kernel, the resulting regularization is inherently bounded and avoids excessive variance. From a theoretical standpoint, this mixture Gaussian regularization directly shapes the policy update through its adaptive, smoothly decaying penalty:
\begin{equation}
L_{\mathrm{MixGaussian}} = \frac{1}{K} \sum_{i=1}^{K} \left( \frac{\pi_i(a \mid s)}{\mu(a \mid s)} \cdot \frac{1}{2} \left[ 1 - \exp\left( - \left( \frac{\pi_\theta(a \mid s)}{\pi_i(a \mid s)} - 1 \right)^2 \right) \right] \right)
\end{equation}
where $\mu$ denotes the behavior policy, $\pi_i$ represents the policy at the $i$-th epoch (with $\pi_1 = \mu$), and $\pi_\theta$ is the current policy. The final \texttt{GTR} objective is thus given by:
\begin{equation}
L_{\mathrm{GTR}} = \mathbb{E}_{t} \left\{ - \min \left[ \frac{\pi_\theta(a \mid s)}{\mu(a \mid s)} \hat{A}_t, \operatorname{clip}\left(\frac{\pi_\theta(a \mid s)}{\mu(a \mid s)}, 1 - \epsilon, 1 + \epsilon\right) \hat{A}_t \right] + \eta L_{\mathrm{MixGaussian}} \right\}
\end{equation}
\paragraph{Rethink the Rationality of \texttt{GTR} According to Fisher Information Matrix}
From an optimization perspective, we construct a trust region along \emph{the path of policy improvement}~\citep{dabney2021value}. From a gradient perspective, the probability measures of reference policies are unified to the current parameterized policy $\pi_\theta$ via the Radon-Nikodym derivative (i.e. the importance weight). Concurrently, a Gaussian kernel is applied to these derivatives to smoothly penalize deviations, yielding:
\begin{equation}
    \nabla_\theta L_{\mathrm{MixGaussian}} = \mathbb{E}_{\pi_\theta} \Biggl[ \frac{1}{K} \sum_{i=1}^{K} \underbrace{\exp\left( - \left( \frac{\pi_\theta(a \mid s)}{\pi_i(a \mid s)} - 1 \right)^2 \right)}_{\text{\scriptsize Gaussian Kernel}} \cdot \underbrace{\left( \frac{\pi_\theta(a \mid s)}{\pi_i(a \mid s)} - 1 \right)}_{\text{\scriptsize Trust Region Penalty}} \nabla_\theta \log \pi_\theta \Biggr]
\end{equation}
Our approach can thus be viewed as approximating the effective FIM via a Gaussian-weighted aggregation of multiple local FIMs along the optimization trajectory.
\section{Experiments}\label{sec:4}
We evaluate \texttt{GTR} across a diverse spectrum of architectures and tasks ranging in complexity, to determine whether continual learning ability is effectively recovered. Our evaluation spans three distinct domains: (i) Robotic Control: Evaluating continual learning performance and generalization across tasks. (ii) Open-World Exploration: Assessing adaptability under prolonged training with high data staleness. (iii) Language Reasoning Model Post-Training: Demonstrating the efficacy of \texttt{GTR} on large-scale models, highlighting the mode transition issue in such domain. (iv) Ablation Study: Investigating the robustness of \texttt{GTR} to relaxed clipping and examining whether adjusting KL's penalty strength can replicate the effect of \texttt{GTR}.
\subsection{Robotic Control: \texttt{GTR} Stimulates Continual Learning Capacity with Advanced Architecture}
\paragraph{Experimental Setup}

We compare \texttt{GTR} with standard PPO, PPO with a KL-based trust-region constraint and a recent PPO variant, SPO~\citep{xie2024simple}. All results are averaged over three random seeds, and each method is tuned with empirically optimal hyperparameters (details in \autoref{hyper}). We adopt the same SimBa~\citep{lee2024simba} architecture (advanced scalable residual blocks with LayerNorm) to parameterize policies to ensure baseline performance and eliminate the performance degradation caused by network expressiveness. To comprehensively evaluate continual learning ability, we construct two types of benchmarks:

\paragraph{Sequential Training on Related Tasks within the Same Scenario} This setting evaluates the ability to adapt between tasks with shared action and observation spaces (use the same embodied skeleton) but differing task goals, capturing smooth behavioral mode transitions. We build two difficult locomotion benchmarks via DeepMind Control Suite~\citep{tassa2018deepmind}: (i) Walker: \textit{walker stand$\to$walk$\to$run} and (ii) Dog: \textit{dog stand$\to$walk$\to$run$\to$trot}, repeated for two cycles.

\begin{wrapfigure}{r}{0.54\textwidth}
\centering
\vspace{-0.6cm}
    \includegraphics[width=\linewidth]{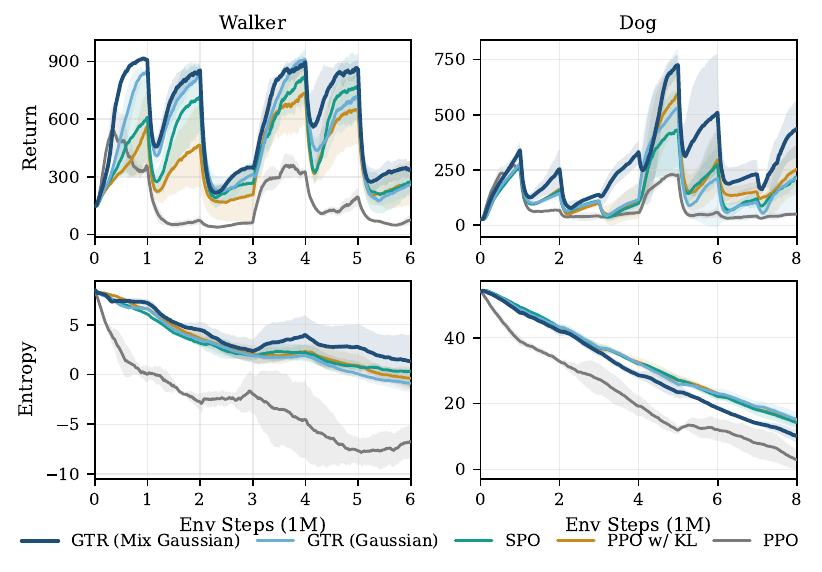}
    \caption{\textbf{Performance and policy entropy with the default Simba architecture.} (Top row) Episode return across two benchmarks. (Bottom row) Corresponding
policy entropy. Results are averaged over three seeds.}
    \label{fig:exp_1}
    \vspace{-0.6cm}
\end{wrapfigure}\paragraph{Sequential Training on Differentiated Tasks across Scenarios} This setting evaluates adaptation under significant distribution shifts, including differences in task logic, observation spaces, and action spaces. It poses a more challenging test of policy plasticity. We construct two benchmarks using MuJoCo-v5 tasks \citep{todorov2012mujoco}, \textit{HalfCheetah$\to$Walker$\to$Ant$\to$Humanoid}, along with the reversed order, trained sequentially over three cycles.
\paragraph{Results: Related Tasks Sequential Training} As shown in \autoref{fig:exp_1} (first row), \texttt{GTR} consistently achieves the best performance across all benchmarks and outperforms strong baselines by up to $25\%$ in peak performance. \texttt{GTR} exhibits substantially faster adaptation when switching tasks, as evidenced by the rapid rise in performance immediately after task transitions. This indicates more efficient behavioral mode switching. Further analysis of policy entropy (\autoref{fig:exp_1} (second row)) shows that \texttt{GTR} maintains higher entropy during training. This suggests that the reshaped trust region preserves exploration capacity and enables efficient mode seeking in a huge solution space.
\begin{figure}[h!]
    \centering
    \includegraphics[width=1.0\linewidth]{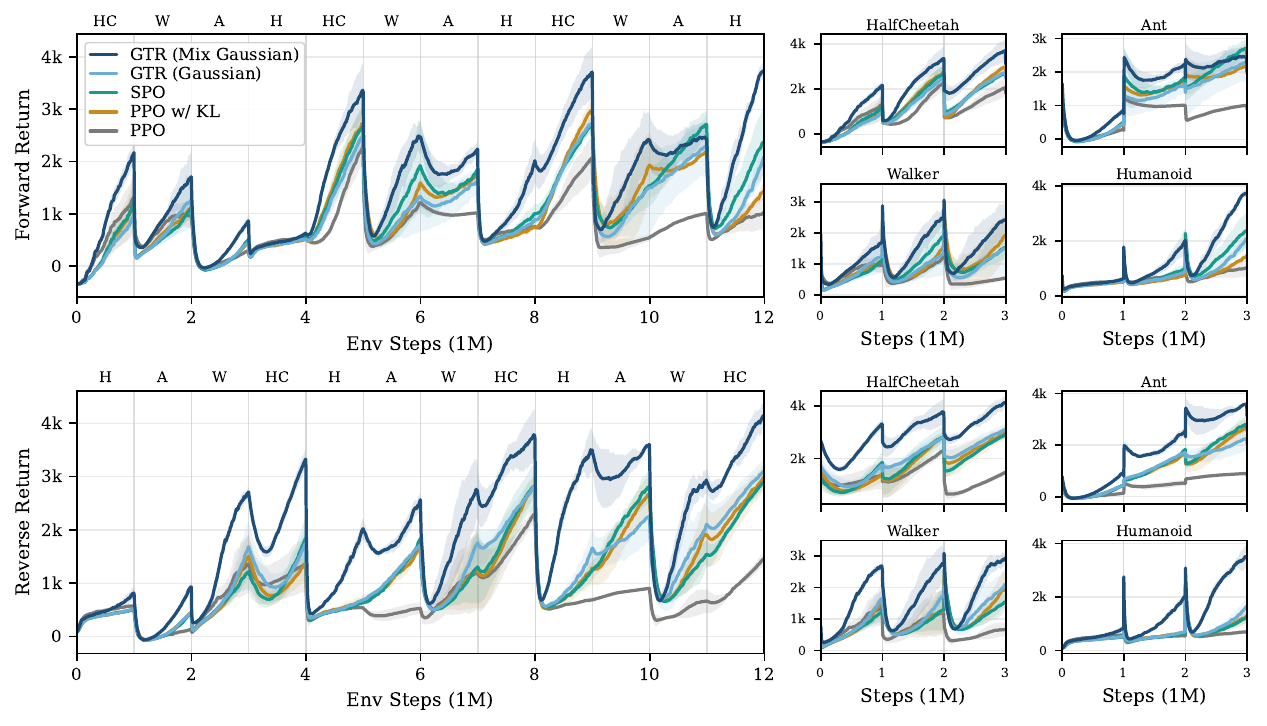}
    \caption{ (Top row): \textbf{Episode return during forward loop training across four tasks}, i.e., H: Humanoid, A: Ant, W: Walk, HC: HalfCheetah. (Bottom row) \textbf{Episode return during inverse loop training.} \texttt{GTR} consistently achieves best performance during each cycle.}
    \label{fig:exp_2}
    \vspace{-0.1cm}
\end{figure}
\paragraph{Results: Differentiated Tasks Sequential Training} When task differences become more significant, the benefits of \texttt{GTR} are further amplified. Comparing \autoref{fig:exp_2} (left) with Figure~1, we observe that increased difficulty in behavior mode transitions leads to larger performance gaps between \texttt{GTR} and baseline methods. \texttt{GTR} consistently achieves superior performance under both forward and reverse training orders, demonstrating robustness to curriculum variations. This indicates that the method is less sensitive to task ordering and does not rely on favorable training sequences. A per-task analysis (\autoref{fig:exp_2} (right)) further reveals that the mixture Gaussian trust region enables more precise modeling of behavioral boundaries. As a result, performance improves progressively across training cycles, with most tasks surpassing their previous best scores. Since state and action spaces are different across tasks, we employ task-specific input layers for both actors and critics and task-specific output heads for actors. Such layers account for a small fraction of the total parameters.
\subsection{Open-World Exploration: \texttt{GTR} Discovers and Learns New Skills during Prolonged Training}
\begin{figure}[h]
    \centering
    \includegraphics[width=1.0\linewidth]{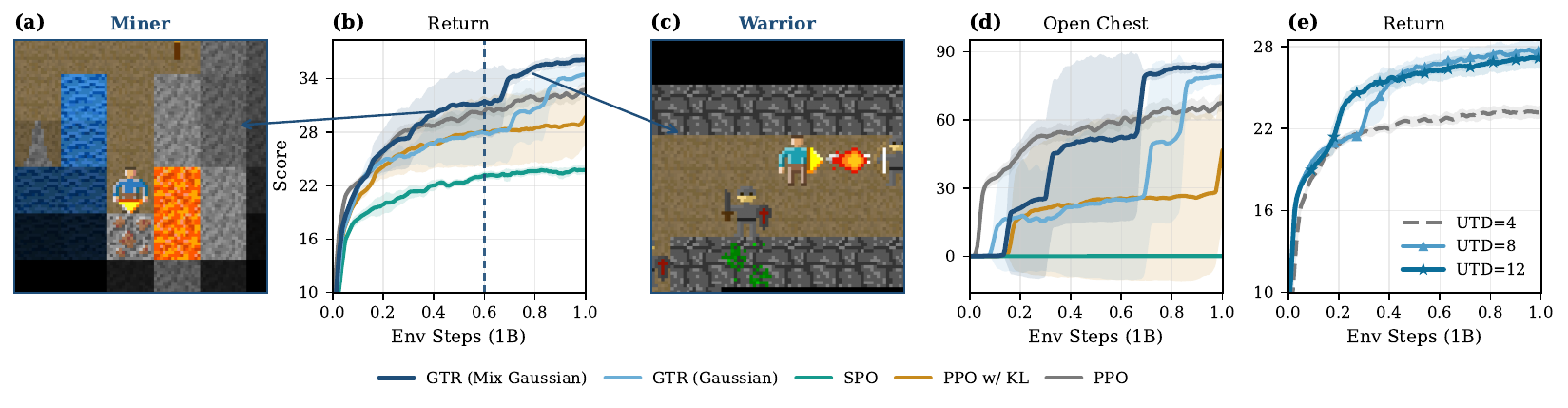}
    \caption{(a-c): \textbf{Episodic return of GRU-based PPO}, where arrows illustrate the agent's mode shift from miner to warrior (d): \textbf{Performance of GRU-based PPO on \textit{open-chest}} (e): \textbf{Ablation study on the Update-to-Data ratio of SimBa-based PPO.} Results are averaged over three seeds.}
    \label{fig:exp_3}
    \vspace{-0.2cm}
\end{figure}\paragraph{Experimental Setup} To evaluate whether \texttt{GTR} can achieve obvious behavior mode transition and stable training in prolonged training, we conduct an extremely long-horizon experiment in a challenging Minecraft-style open-world environment, Craftax~\citep{matthews2024craftax}, which features a rich set of compositional subtasks and diverse behavioral patterns. We consider two architectures: (i) SimBa, an advanced residual network, following the official implementation and (ii) a GRU backbone with LayerNorm~\citep{ba2016layer,nauman2024overestimation}, following the implementation in \citet{matthews2024craftax}. Agents are trained for $1$ billion steps to test the ability of behavior transition.

\begin{wrapfigure}{r}{0.44\textwidth}
\centering
\vspace{-0.4cm}
    \includegraphics[width=\linewidth]{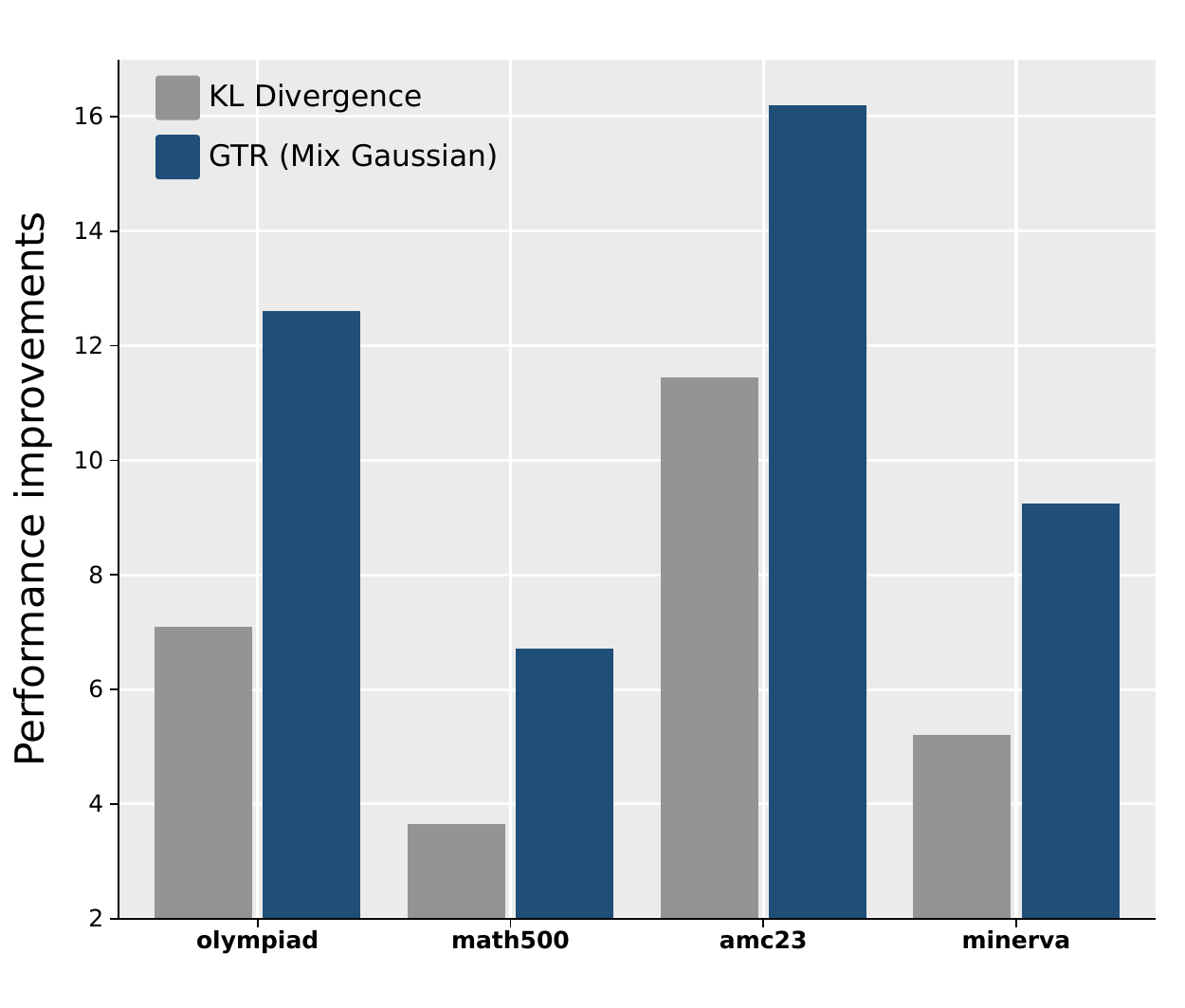}
    \vspace{-0.6cm}
    \caption{\textbf{Score Improvement over the initial policy on four tasks.} Both methods use GRPO as a backbone.}
    \label{llm}
    \vspace{-0.4cm}
\end{wrapfigure}\paragraph{Results}
In experiments with GRU-based PPO (\autoref{fig:exp_3}(b)), we identify \textit{open-chest} (recorded in \autoref{fig:exp_3}(d)) as a critical milestone highly predictive of ultimate performance. Prior to mastering this skill, the agent defaults to a conservative mining strategy (termed the \emph{old behavior}). Upon acquiring \textit{open-chest}, the agent secures sufficient resources to hunt formidable monsters, effectively transitioning into a ``warrior'' (the \emph{new behavior}). We observe an intriguing dynamic: while standard PPO discovers \textit{open-chest} early in training, accumulated inefficiencies in local optimization leave it entrenched in the old pattern. SPO, hampered by its aggressive regularization against outliers, completely fails to transition into the new pattern, converging to a suboptimal policy. Conversely, although \texttt{GTR} learns \textit{open-chest} slightly later in the training process, its exceptional learning capacity enables a rapid behavioral shift into the new mode, ultimately yielding the highest return.

We further investigate performance under high update-to-data (UTD) regimes using SimBa-based PPO (as shown in \autoref{fig:exp_3}(e)). As we increase the number of PPO epochs from 4 to 12, \texttt{GTR} not only maintains stable training, but also learns \textit{open-chest} significantly faster and converges to higher returns. This suggests that \texttt{GTR} effectively leverages high-UTD updates to accelerate adaptation.

\subsection{LLM Training: \texttt{GTR} Reveals the Implicit Need for Mode Transition in Fine-tuning}

We adopt GRPO~\citep{guo2025deepseek}, a simplified variant of PPO widely used in LLM post-training, as the basis, where group normalization replaces GAE for advantage estimation. We set $UTD=2$ (equivalent to PPO epochs), introducing mild off-policy effects due to data reuse. Policies are initialized from \texttt{Qwen3-1.7B-Base}. To ensure reproducibility and fairness, we exclusively use open-source datasets in \citet{yu2025dapo}.  We evaluate performance using pass@4 on four challenging benchmarks: MATH-500~\citep{lightman2023let}, OlympiadBench~\citep{he2024olympiadbench}, MinervaMath~\citep{lewkowycz2022solving}, and AMC 2023~\citep{xue2025simpletir}. See \autoref{app:llm} for detailed settings.
\paragraph{Results} As shown in \autoref{llm}, \texttt{GTR} achieves the best performance across all benchmarks. We attribute this improvement to the implicit continual learning nature of LLM post-training. In each training batch, prompts correspond to diverse reasoning prompts, requiring the policy to continuously adapt across heterogeneous objectives while maintaining previously acquired capabilities. As a result, optimization dynamics naturally involve frequent shifts between different behavioral patterns, and local directional guidance plays a critical role in achieving strong reasoning performance.

\subsection{Ablation study}
\begin{wrapfigure}{r}{0.6\textwidth}
    \centering
    \vspace{-0.8cm}
    \includegraphics[width=\linewidth]{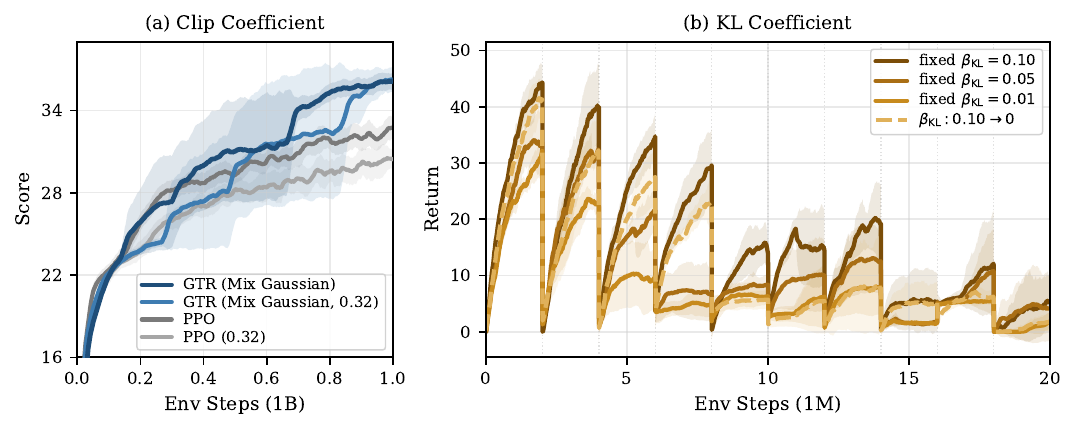}
    \caption{(Left): \texttt{GTR} remains robust on large clip range. (Right): PPO still collapses when adjusting KL coefficient.}
    \label{ablation}
    \vspace{-0.6cm}
\end{wrapfigure}
\paragraph{\texttt{GTR} shows robustness to relaxed clipping} As illustrated in \autoref{ablation} (Left), after increasing the clip range of GRU-based PPO in Craftax, although the convergence slows down, it does not affect the final performance of \texttt{GTR}. In contrast, PPO experiences performance degradation. This demonstrates that the Gaussian trust region is more reliable to guide the local landscape.
\paragraph{Modifying KL strength fails to recover the learning ability of standard PPO} As depicted in \autoref{ablation} (Right), we evaluated three fixed KL coefficients alongside a linearly decaying KL schedule starting from $0.1$. All configurations led to collapse in Procgen. This highlights the necessity of directly reshaping the trust region to unlock the continual learning ability.

\section{Related Work}
While stable dynamics and signals are theoretically crucial for algorithm convergence in reinforcement learning~\citep{agarwal2021theory}, numerous empirical practices in model-free RL have demonstrated the robustness of deep RL to such tasks, utilizing techniques like residual learning~\citep{silver2018residual} and network ensembling~\citep{ball2023efficient}. Interestingly, these methods are entirely based on Q-learning~\citep{watkins1992q}. Rather than contradicting foundational theories, these empirical techniques effectively stabilize the variance and mitigate non-stationary, thereby satisfying the implicit requirements of classic stochastic approximation theory~\citep{borkar2008stochastic} and Robbins-Monro theorem~\citep{robbins1951stochastic} in highly non-linear settings. Furthermore, the replay buffer mechanism in Q-learning naturally aligns with classic continual learning algorithms, such as EWC~\citep{kirkpatrick2017overcoming} and PackNet~\citep{mallya2018packnet}, which has directly shaped the mainstream research trajectory of continual RL in recent years.

Compared to off-policy counterparts, research on continual learning based on PPO remains exceptionally scarce. Prior mainstream research has undergone a paradigm shift from learning dynamics~\citep{lyle2022learning,tang2023towards}, generalization and interference~\citep{bengio2020interference}, towards plasticity~\citep{dohare2024loss}. The latter specifically emphasizes the plasticity of neural networks, and existing methods have explored this from a diverse array of perspectives: plasticity injection~\citep{nikishin2023deep}, loss landscapes~\citep{lyle2023understanding,castanyer2026stable,pasand2026stable}, regularization~\citep{lyle2024disentangling,obandoceron2025}, network weights~\citep{elsayed2024weight}, effective learning rates~\citep{lyle2024normalization}, dormant neurons~\citep{sokar2023dormant}, the lottery ticket hypothesis~\citep{graesser2022statesparsetrainingdeep,ceron2024in,liu2024neuroplastic}, gradient geometry~\citep{liu2025measure}, and Neural Tangent Kernel (NTK) theory~\citep{tang2025mitigating}. While fruitful, these approaches often blur the mechanistic distinctions between policy gradient methods and other RL algorithms, leaving the inherent continual learning potential of PPO itself largely overlooked.

Although our foundational work builds upon vanilla PPO, numerous exceptional variants have been derived from PPO at the application level, which may harbor even greater potential for continual learning. PPO-EWMA~\citep{hilton2022batch} investigates asynchronous settings under limited computational resources, PPG~\citep{cobbe2021phasic} explores representation-level sharing and data efficiency, and GRPO~\citep{guo2025deepseek} has successfully enabled the RL training of reasoning models. Additionally, the data collection pipeline~\citep{mayor2025impact} itself warrants further consideration. We leave the exploration of these promising directions for future work.

\section{Conclusion}

In this work, we investigated the limitations of PPO in continual and non-stationary learning settings. Through systematic empirical analysis, we found that policy collapse is associated less with insufficient model capacity or clip range and more with ineffective local exploration. Without geometry-aware guidance, policy updates may become directionally misaligned, leading to the accumulation of biased local updates that hinder global adaptation. While divergence-based constraints introduce partial geometric structure, their monotonically increasing penalties may restrict transitions toward emerging behavior modes.

To address this limitation, we proposed \texttt{GTR}, a trust-region formulation based on a Gaussian kernel that balances local stability with controlled exploration. By further introducing a mixture-based anchor that adapts to evolving policies, \texttt{GTR} enables efficient local optimization while facilitating gradual and reliable behavioral transitions. Extensive experiments across diverse domains show that \texttt{GTR} consistently improves performance over PPO baselines, supporting the importance of trust-region geometry in non-stationary learning. We hope this work motivates further research on trust-region design as a foundation for more robust and adaptive reinforcement learning.

\textbf{Limitations.} First, due to computational constraints, we do not evaluate \texttt{GTR} in large-scale language model training or across varying LLM sizes; extending geometry-aware trust-region methods to such high-compute regimes remains an important direction for future work. Second, although the Gaussian-shaped trust-region formulation is empirically effective and well motivated, its design remains primarily grounded in empirical observations and intuition rather than a complete theoretical characterization. Developing a deeper theoretical understanding of geometry-aware and non-monotonic regularization is an important avenue for future research.

\clearpage
\bibliography{main}
\bibliographystyle{plainnat}

\appendix

\section{Extended Preliminaries and Derivations}\label{theo}
\subsection{Divergence and Fisher Information}

Let $\pi_\theta(a \mid s)$ be a differentiable policy, and consider a local perturbation $\theta' = \theta + \Delta \theta$. The forward KL divergence between $\pi_\theta$ and $\pi_{\theta'}$ is defined as:
\begin{equation}
D_{\mathrm{KL}}(\pi_\theta \,\|\, \pi_{\theta'}) 
= \mathbb{E}_{s \sim d^{\pi_\theta},\, a \sim \pi_\theta}
\left[
\log \frac{\pi_\theta(a \mid s)}{\pi_{\theta'}(a \mid s)}
\right].
\end{equation}

\paragraph{Proposition 1 (Second-Order Expansion of KL and FIM).}
For sufficiently small $\Delta \theta$, the KL divergence admits the following expansion:
\begin{equation}
D_{\mathrm{KL}}(\pi_\theta \,\|\, \pi_{\theta + \Delta \theta})
= \tfrac{1}{2}\,\Delta \theta^\top F(\theta)\,\Delta \theta 
+ o(\|\Delta \theta\|^2),
\end{equation}
where
\begin{equation}
F(\theta) 
= \mathbb{E}_{s \sim d^{\pi_\theta},\, a \sim \pi_\theta}
\left[
\nabla_\theta \log \pi_\theta(a \mid s)\,
\nabla_\theta \log \pi_\theta(a \mid s)^\top
\right]
\end{equation}
is the Fisher Information Matrix (FIM).

\paragraph{Proof (Sketch).}
Let $\ell(\theta') := \log \pi_{\theta'}(a \mid s)$. Applying a second-order Taylor expansion of $\ell(\theta')$ around $\theta$, we obtain:
\begin{equation}
\ell(\theta + \Delta \theta)
= \ell(\theta)
+ \nabla_\theta \ell(\theta)^\top \Delta \theta
+ \tfrac{1}{2}\Delta \theta^\top \nabla_\theta^2 \ell(\theta)\,\Delta \theta
+ o(\|\Delta \theta\|^2).
\end{equation}

Substituting into the KL definition:
\begin{equation}
D_{\mathrm{KL}}(\pi_\theta \,\|\, \pi_{\theta + \Delta \theta})
= \mathbb{E}\big[\ell(\theta) - \ell(\theta + \Delta \theta)\big].
\end{equation}

Taking expectations:
\begin{itemize}
    \item The first-order term vanishes:
    \begin{equation}
    \mathbb{E}_{a \sim \pi_\theta}
    \big[\nabla_\theta \log \pi_\theta(a \mid s)\big] = 0.
    \end{equation}

    \item The second-order term becomes:
    \begin{equation}
    - \tfrac{1}{2}\Delta \theta^\top 
    \mathbb{E}\big[\nabla_\theta^2 \log \pi_\theta(a \mid s)\big]
    \Delta \theta.
    \end{equation}
\end{itemize}

Using the identity:
\begin{equation}
F(\theta) = - \mathbb{E}\big[\nabla_\theta^2 \log \pi_\theta(a \mid s)\big],
\end{equation}
we obtain:
\begin{equation}
D_{\mathrm{KL}}(\pi_\theta \,\|\, \pi_{\theta + \Delta \theta})
= \tfrac{1}{2}\Delta \theta^\top F(\theta)\,\Delta \theta
+ o(\|\Delta \theta\|^2).
\end{equation}
\hfill $\square$

\paragraph{Remarks.}
KL divergence locally induces a quadratic form defined by the FIM, thereby providing a geometry-aware metric over the policy space. Consequently, KL-based trust-region methods implicitly constrain updates under the Fisher geometry for small policy perturbations.

\subsection{Derivation of Score Function Penalties}\label{derivegaussian}

Let $\tilde{r} = \frac{\pi_\theta(a|s)}{\pi_{\mathrm{old}}(a|s)}$ denote the probability ratio. Using the identity $\nabla_\theta \tilde{r} = \tilde{r} \nabla_\theta \log \pi_\theta$, a general gradient penalty of the form $\nabla_\theta L = \mathbb{E}_{\pi_{\mathrm{old}}} [\lambda(\tilde{r}) \nabla_\theta \log \pi_\theta]$ corresponds to a pointwise penalty function $f(\tilde{r})$ satisfying:
\begin{equation}
    \mathbb{E}_{\pi_{\mathrm{old}}} \left[ f'(\tilde{r}) \nabla_\theta \tilde{r} \right] = \mathbb{E}_{\pi_{\mathrm{old}}} \left[ \underbrace{\tilde{r} f'(\tilde{r})}_{:= \lambda(\tilde{r})} \nabla_\theta \log \pi_\theta \right].
\end{equation}
This yields the general integration rule:
\begin{equation}
    f(\tilde{r}) = \int \frac{\lambda(\tilde{r})}{\tilde{r}} \, d\tilde{r}.
\end{equation}

Applying this rule, we derive three specific penalty objectives based on the choice of the score function multiplier $\lambda(\tilde{r})$:

\textbf{Inverse KL divergence:} Let $\lambda(\tilde{r}) = \tilde{r} \log \tilde{r}$. Integrating yields:
\begin{equation}
    f(\tilde{r}) = \int \frac{\tilde{r} \log \tilde{r}}{\tilde{r}} \, d\tilde{r} = \int \log \tilde{r} \, d\tilde{r} = \tilde{r} \log \tilde{r} - \tilde{r} + 1.
\end{equation}

\textbf{Pearson $\chi^2$-divergence:} Let $\lambda(\tilde{r}) = \tilde{r}(\tilde{r} - 1)$. Integrating yields:
\begin{equation}
    f(\tilde{r}) = \int \frac{\tilde{r}(\tilde{r} - 1)}{\tilde{r}} \, d\tilde{r} = \int (\tilde{r} - 1) \, d\tilde{r} = \frac{1}{2}(\tilde{r} - 1)^2.
\end{equation}

\textbf{Gaussian penalty:} Let $\lambda(\tilde{r}) = \tilde{r}(\tilde{r} - 1)\exp(-(\tilde{r} - 1)^2)$. Integrating yields:
\begin{equation}
    f(\tilde{r}) = \int (\tilde{r} - 1)\exp\left(-(\tilde{r} - 1)^2\right) \, d\tilde{r} = \frac{1}{2}\left[ 1 - \exp\left(-(\tilde{r} - 1)^2\right) \right].
\end{equation}

\section{Hyper-parameter setting}\label{hyper}
\subsection{LLM Post-training}\label{app:llm}
To ensure consistency with prior work, we fix the global batch size to $1024$ and the maximum response length to $8192$ tokens. We use VERL \citep{sheng2025hybridflow} as the infrastructure. The learning rate is set to $1\times10^{-6}$. We adopt GRPO~\citep{guo2025deepseek}—a simplified PPO variant widely used in LLM post-training—as the optimization backbone, where group normalization replaces GAE for advantage estimation. For text generation, we use nucleus sampling with $\text{top}_p=0.99$, $\text{top}_k=100$, and temperature $0.6$. Each task was trained for 500 steps. We set the update-to-data ratio to $UTD=2$ (equivalent to PPO epochs), introducing mild off-policy effects due to data reuse. The policy is initialized from \texttt{Qwen3-1.7B-Base}. We exclusively use open-source datasets. We follow the training data from \citet{yu2025dapo}, which is known to expose clear performance gaps across algorithms on long-tail reasoning tasks. We evaluate performance using pass@4 on four challenging benchmarks: MATH-500~\citep{lightman2023let}, OlympiadBench~\citep{he2024olympiadbench}, MinervaMath~\citep{lewkowycz2022solving}, and AMC 2023~\citep{xue2025simpletir}.
\subsection{Robotic Control and Open-World Exploration}
For tasks in these two domains, we employ SimBa and RNN architectures, SimBa~\citep{lee2024simba} is a residual neural network tailored specifically for RL that incorporates a simplicity bias. Its efficacy has been well-validated in the original paper across environments such as DeepMind Control Suite~\citep{tassa2018deepmind}, MuJoCo~\citep{todorov2012mujoco}, and Craftax~\citep{matthews2024craftax}. For RNN baseline, we adopt Recurrent-PPO~\citep{akkaya2019solving,berner2019dota}, a widely used PPO variant. To minimize the confounding effects of network plasticity~\citep{lyle2024disentangling,nauman2024overestimation}, we introduce LayerNorm into the official implementation~\citep{matthews2024craftax}.

Furthermore, to explicitly evaluate the model's continual learning capabilities, we maintain a constant learning rate throughout training rather than applying learning rate annealing. Unlike the LLM setup, we employ advantage-weighted penalty, formulated as $\eta = \frac{|A|}{\epsilon}$. Empirically, this dynamic formulation yields superior performance on control tasks compared to utilizing a fixed penalty coefficient~\citep{xie2024simple}. Aside from these specific modifications, we strictly adhere to the official hyperparameter configurations. All experiments were conducted on NVIDIA GeForce RTX 3090 GPU and completed within 12 hours.

\begin{table}[htbp]
\centering
\caption{SimBaBlock structure}
\label{tab:simba_block}
\begin{tabular}{ll}
\toprule
Step & Layer Configuration \\
\midrule
1. Norm & LayerNorm($H$) \\
2. FC + Act & Linear($H, 4H$), ReLU \\
3. FC & Linear($4H, H$) \\
4. Residual & Output = Step 3 Output + Block Input \\
\bottomrule
\end{tabular}
\end{table}

\begin{table}[htbp]
\centering
\caption{SimBa actor-critic network architectures}
\label{tab:dmc_simba_architecture}
\begin{tabular}{lll}
\toprule
Layer & Actor Network & Critic Network \\
\midrule
Fully Connected & Linear($d_{\mathrm{obs}}, 128$) & Linear($d_{\mathrm{obs}}, 512$) \\
SimBaBlock & $1 \times$ SimBaBlock & $2 \times$ SimBaBlock \\
LayerNorm & LayerNorm(128) & LayerNorm(512) \\
Output Head & Linear($128, d_{\mathrm{act}}$) & Linear($512, 1$) \\
Policy Std & Learnable log std & -- \\
\bottomrule
\end{tabular}
\end{table}

\begin{table}[htbp]
\centering
\caption{Layer-normalized GRU cell}
\label{tab:ln_rnn_cell}
\begin{tabular}{ll}
\toprule
Step & Layer Configuration \\
\midrule
1. Gates & Linear($x, 2H$) + Linear($h, 2H$) \\
2. Gate Norm & LayerNorm($2H$), sigmoid \\
3. Candidate & Linear($x, H$) + Linear($r \odot h, H$) \\
4. Candidate Norm & LayerNorm($H$), tanh \\
5. Hidden Update & $h' = (1 - z) \odot \tilde{h} + z \odot h$ \\
\bottomrule
\end{tabular}
\end{table}

\begin{table}[h]
\centering
\caption{LN-RNN actor-critic architecture for Craftax, $H=512$ by default}
\label{tab:craftax_ln_rnn_architecture}
\begin{tabular}{lll}
\toprule
Layer & Actor Network & Critic Network \\
\midrule
Input Projection & \multicolumn{2}{c}{Linear($d_{\mathrm{obs}}, H$), LayerNorm($H$), ReLU} \\
Recurrent Core & \multicolumn{2}{c}{LayerNormGRUCell($H$)} \\
Fully Connected & Linear($H, H$), LayerNorm, ReLU & Linear($H, H$), LayerNorm, ReLU \\
Fully Connected & Linear($H, H$), LayerNorm, ReLU & Linear($H, H$), LayerNorm, ReLU \\
Output Head & Linear($H, d_{\mathrm{act}}$) & Linear($H, 1$) \\
Policy & Categorical logits & -- \\
\bottomrule
\end{tabular}
\end{table}

\begin{table}[htbp]
\centering
\caption{PPO hyperparameter settings across the three domains}
\label{tab:ppo_hyperparameters_all_domains}
\begin{tabular}{lccc}
\toprule
Hyperparameter & DMC & MuJoCo-v5 & Craftax \\
\midrule
Learning Rate & $3 \times 10^{-4}$ & $3 \times 10^{-4}$ & $2 \times 10^{-4}$ \\
Optimizer & Adam & Adam & Adam \\
Optimizer $\epsilon$ & $1 \times 10^{-5}$ & $1 \times 10^{-5}$ & $1 \times 10^{-5}$ \\
Discount Factor ($\gamma$) & 0.99 & 0.99 & 0.99 \\
GAE Parameter ($\lambda$) & 0.95 & 0.95 & 0.8 \\
Number of Environments & 1 & 1 & 1024 \\
Rollout Length & 2048 & 2048 & 64 \\
Batch Size & 2048 & 2048 & 65536 \\
Minibatches & 32 & 32 & 8 \\
Minibatch Size & 64 & 64 & 8192 \\
Update Epochs & 10 & 10 & 4 \\
PPO Clip Coefficient & 0.2 & 0.2 & 0.2 \\
Value Loss Clipping & True & True & True \\
Value Loss Coefficient & 0.5 & 0.5 & 0.5 \\
Entropy Coefficient & 0.0 & 0.0 & 0.01 \\
Max Gradient Norm & 0.5 & 0.5 & 1.0 \\
Advantage Normalization & True & True & True \\
Learning Rate Annealing & False & False & False \\
Training Budget & $1 \times 10^6$ per task & $1 \times 10^6$ per task & $1 \times 10^9$ total \\
\bottomrule
\end{tabular}
\end{table}

\newpage
\section{Additional experiments}
Additionally, we conducted experiments on MetaWorld+~\citep{mclean2025meta} and Procgen~\citep{cobbe2020leveraging}. For MetaWorld+, we utilized tanh-activated MLP on \textit{door open$\to$close} and \textit{faucet open$\to$close}, repeated for three cycles. For Procgen, we utilized the original IMPALA-CNN~\citep{cobbe2020leveraging} on starpilot. As these environments are relatively straightforward, we include the results in this section for completeness and reference.
\begin{figure}[htbp]
    \centering
    \includegraphics[width=0.75\linewidth]{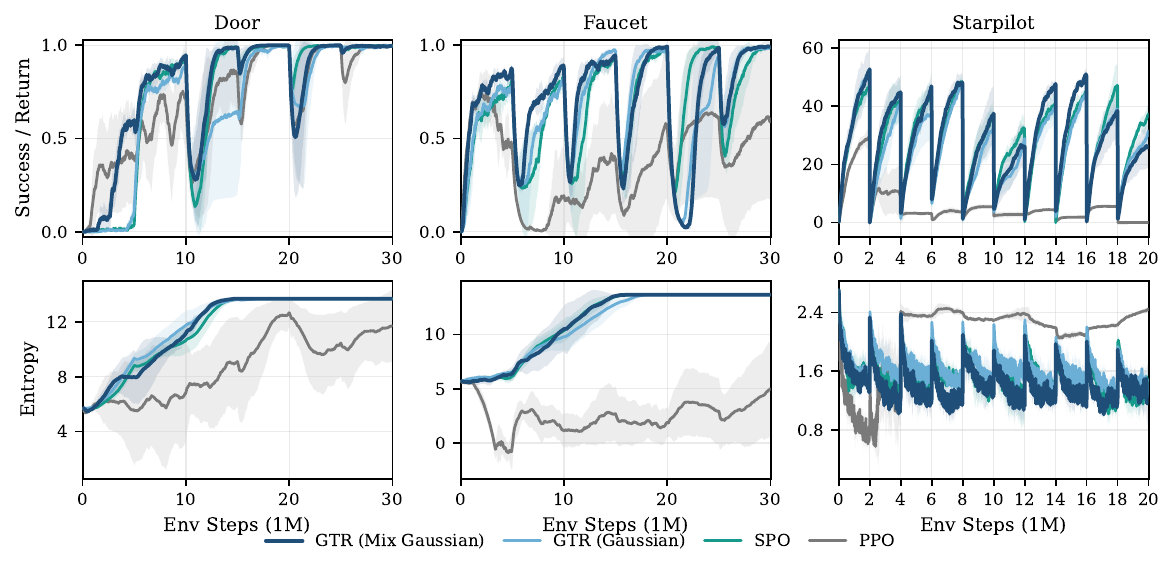}
    \caption{Aside from vanilla PPO, all variants demonstrate superior continual learnability.}
    \label{additional}
\end{figure}
\section{Connect with Student Authors}
% Core Contributor: Bingxu Liu \texttt{bxliu124@gmail.com}. Project leader: Jiashun Liu \texttt{ljshasdream@gmail.com}. 

% Contributor: Johan Obando-Ceron \texttt{jobando0730@gmail.com}, Hao Wang \texttt{hao.wang@my.cityu.edu.hk}, Runze Liu \texttt{1036433215@qq.com}
\noindent
\begin{tabular*}{\textwidth}{@{\extracolsep{\fill}} L{0.56\textwidth} L{0.4\textwidth} @{}}
Bingxu Liu {\normalsize\textup{(Core Contributor)}} & \texttt{bxliu124@gmail.com} \\
Jiashun Liu {\normalsize\textup{(Project Leader)}} & \texttt{ljshasdream@gmail.com} \\
Johan Obando-Ceron & \texttt{jobando0730@gmail.com} \\
Hao Wang & \texttt{hao.wang@my.cityu.edu.hk} \\
Runze Liu & \texttt{1036433215@qq.com}
\end{tabular*}

\end{document}